\documentclass[conference]{IEEEtran}
\IEEEoverridecommandlockouts
\usepackage{cite}
\usepackage{amsmath,amssymb,amsfonts}
\usepackage{algorithmic}
\usepackage{graphicx}
\usepackage{textcomp}

\usepackage{xcolor}
\def\BibTeX{{\rm B\kern-.05em{\sc i\kern-.025em b}\kern-.08em
    T\kern-.1667em\lower.7ex\hbox{E}\kern-.125emX}}

\usepackage{amsthm}

\usepackage{comment}
\usepackage{booktabs} 
\usepackage{graphicx}
\usepackage{subfig}
\usepackage{stfloats}

\begin{document}

\title{What Physics do Data-Driven MoCap-to-Radar Models Learn? 
}


\author{\IEEEauthorblockN{Kevin Chen}
\IEEEauthorblockA{\textit{Computer Science \& Engineering} \\
\textit{The Ohio State University}\\
Columbus, Ohio\\
\underline{chen.11020@osu.edu}}
\and
\IEEEauthorblockN{Kenneth W. Parker}
\IEEEauthorblockA{\textit{Unaffiliated} \\
Waterford, Virginia\\
\underline{kenneth@parkertong.net}}
\and
\IEEEauthorblockN{Anish Arora}
\IEEEauthorblockA{\textit{Computer Science \& Engineering} \\
\textit{The Ohio State University}\\
Columbus, Ohio\\
\underline{arora.9@osu.edu}}
}
\maketitle

\begin{abstract}
Data-driven MoCap-to-radar models generate plausible micro-Doppler spectrograms, but do they actually learn the underlying physics? We introduce a physics-based interpretability framework to answer this question via two proposed complementary metrics: one measures alignment between model predictions and the physics-derived Doppler frequency, while the other tests whether predictions preserve the velocity–frequency relationship under velocity intervention. Both metrics require only MoCap input and model predictions, without access to measured radar data. Experiments across several model architectures reveal that low reconstruction error does not guarantee physical consistency: some, but not all, models achieve low error yet perform poorly on the two physics-based metrics. Further analysis shows that temporal attention is critical for transformer-based models to learn the underlying physics.
\end{abstract}


\section{Introduction}

Micro-Doppler signatures play an essential role in radar-based human sensing, capturing fine-scale motion through Doppler shifts induced by the human body. 
These signatures underpin applications such as activity recognition \cite{Shi2018, Tan2024}, health monitoring \cite{Xu2022}, and edge sensing \cite{roy2021one}, reflecting the fact that they arise from motion-induced modulation of radar returns and thus encode information closely related to the underlying kinematics.

Collecting large-scale micro-Doppler datasets remains costly and 
environment-dependent, which has motivated recent efforts to 
synthesize micro-Doppler signatures. Physics-based approaches 
model the body as a collection of point scatterers or apply 
ray-tracing to articulated body meshes 
\cite{erol2015kinect, singh2018simulation, Vishwakarma2022}. 
Hybrid methods combine machine learning (ML) with physics-based radar simulators: they first extract motion representations (e.g., pose or skeleton sequences) from video or textual descriptions, then feed these into a radar simulation pipeline \cite{Ahuja2021, text2doppler2024}. More recently, the purely 
data-driven MoCap2Radar model \cite{chen2026mocap2radar} has 
demonstrated that high-quality micro-Doppler spectrograms can be 
synthesized directly from 3D MoCap data by formulating the task 
as a windowed sequence-to-sequence learning using a spatio-temporal 
transformer.

Yet, despite producing visually plausible spectrograms and achieving low reconstruction error, the MoCap2Radar model is trained without any explicit radar physics priors. Its reconstruction loss optimizes only for spectrogram-domain similarity and remains agnostic to physics, such as the relationships dictated by radial velocity and Doppler frequency. This raises a fundamental question: {\em Do these ML models actually capture the underlying radar physics, or do they merely fit motion-specific or target-specific spectral patterns in the training data}? This question connects to the broader pursuit of interpretable ML \cite{lipton2017mythosmodelinterpretability, molnar2020interpretable, bereska2024mechanisticinterpretabilityaisafety}, particularly the growing interest in techniques that assess whether models internalize underlying physical laws\cite{iten2020discovering, chen2022comphy, janny2022filtered, liu2023counterfactual}. To the best of our knowledge, such analyses have not been explored for MoCap-to-radar generation, and no existing work provides tools to evaluate the physical consistency of learned radar spectrograms. 


Motivated by this gap, we propose a physics-based interpretability framework for probing the physical behavior of purely data-driven MoCap-to-radar models. Our framework leverages two physics-based reference models grounded in radar scattering and the Doppler effect. By evaluating predicted outputs against these physics reference models via the proposed physics-based metrics, our framework quantifies physical consistency, i.e., how well learned models align with expected physical behavior, and goes beyond reconstruction error to reveal whether predictions stem from genuine physical understanding or statistical shortcuts.

Our contributions are threefold:
\begin{enumerate}
\item We develop two complementary physics-based metrics that form the core of 
our interpretability framework, evaluating whether MoCap-to-radar 
models preserve physical consistency without requiring ground truth 
radar data.
\item We show that physical consistency is not captured by 
reconstruction error: models with comparable Mean Absolute Error 
(MAE) can diverge sharply in physical consistency.
\item Our ablation studies further reveal that temporal attention plays 
a critical role in learning the underlying physics: model 
variants lacking temporal attention fail to achieve physical 
consistency.
\end{enumerate}

The rest of the paper is organized as follows.
Section~\ref{sec:proposed} introduces the physics-based metrics that form the core of our interpretability framework.
Section~\ref{sec:exp-setup} describes the experimental settings. Section~\ref{sec:exp-results} presents and discusses the results.
Finally, Section~\ref{sec:conclusion} concludes the paper.


\section{Proposed Interpretability Framework}
\label{sec:proposed}

\subsection{Overview}
\label{sec:overview}
\begin{figure}[!t]
    \centering
    \includegraphics[width=0.8\linewidth]{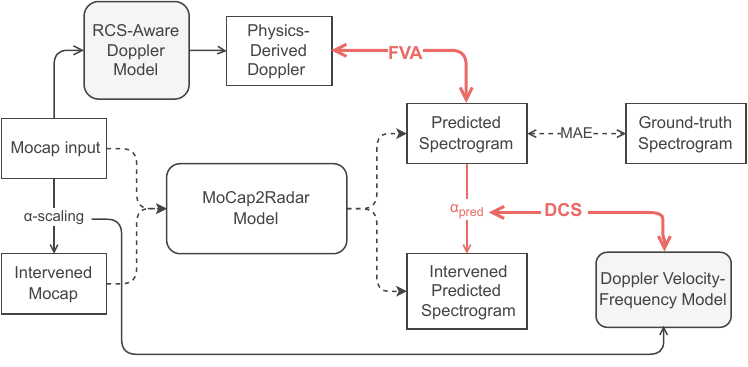}
    \caption{Overview of the interpretability framework, illustrating how the model outputs are processed and analyzed using the proposed metrics.}
    \label{fig:purpose}
\end{figure}

We propose a physics-based interpretability framework for evaluating whether learned MoCap-to-radar models preserve underlying physics. As illustrated in Fig.~\ref{fig:purpose}, rather than relying solely on reconstruction error (MAE), we evaluate the model by comparing its behavior against \emph{two physics-based reference models} grounded in radar physics. The first model, the \emph{RCS-Aware Doppler Model} derives the centroid of the Doppler frequency from the aggregated response of body segments associated with the MoCap markers, whereas the second model, the \emph{Doppler Velocity-Frequency Model}, is based on the linear relationship between radial velocity and Doppler frequency for all MoCap markers, which entails a scaling behavior under velocity intervention.
The proposed metrics quantify whether the learned model's predictions align with these physical expectations.

\subsection{Physics-based Reference Models}
\subsubsection*{RCS-Aware Doppler Model}
This model derives the reference Doppler-centroid trajectory directly from MoCap data, representing the expected Doppler-centroid evolution given the recorded motion. This reference serves as the physical ground truth for evaluating the alignment metric defined in Section~\ref{sec:fva}. The derivation proceeds as follows.

Given a MoCap sequence, let $\mathbf{x}_m(t)\in\mathbb{R}^3$ denote the 3D position of marker $m$ at sample index $t$, upsampled to the radar sampling rate $f_s$. Let $\mathbf{x}_{\mathrm{radar}}$ denote the radar location obtained from the averaged positions of radar-mounted markers, expressed in the same MoCap coordinate frame. The instantaneous range of marker $m$ is
\[
r_m(t) = \|\mathbf{x}_m(t)-\mathbf{x}_{\mathrm{radar}}\|_2,
\]
and the radial velocity is approximated using a central finite difference:
\[
v_m(t) \approx \frac{r_m(t+1)-r_m(t-1)}{2\Delta t},
\qquad \Delta t = 1/f_s.
\]
The corresponding Doppler frequency is then given by:
\begin{equation}
\label{eq:vel-feq}
f_{m}(t) = \frac{2 v_m(t)}{\lambda}.
\end{equation}

Marker-level Doppler frequencies are aggregated using weights $w_m$ based on body surface area (BSA) proportions~\cite{wallace1951burns}, which approximate the relative radar cross-section (RCS) contribution of each body part:
\[
f_c^{\mathrm{ref}}(t)
= \sum_{m=1}^M 
    \frac{w_m}{\sum_j w_j}\, f_{m}(t).
\]
This weighted average prioritizes larger body segments over smaller extremities, reflecting the physical dominance of high-RCS scatterers in radar returns. While BSA serves as only a coarse RCS proxy, it provides a simple, anatomically grounded weighting sufficient for our interpretability analysis.

To match the  short-time Fourier transform (STFT) framing used for spectrogram prediction, the frame-synchronized reference centroid is obtained by averaging over the corresponding window:
\[
f_c^{\mathrm{ref}}[n]
= \frac{1}{L}\sum_{t=nH}^{nH+L-1} f_c^{\mathrm{ref}}(t),
\]
where $L$ is the window length, $H$ is the hop size, and $n$ indexes the STFT frames. This trajectory specifies how Doppler centroids should evolve given the recorded MoCap input.

\subsubsection*{Doppler Velocity-Frequency Model}

This model specifies the expected scaling relationship between radial velocity intervention and Doppler frequency response, which serves as the basis for the metric defined in Section~\ref{sec:dcs}.

Since the Doppler frequency~\eqref{eq:vel-feq} scales linearly with velocity and holds independently for each marker, uniformly scaling all marker 
radial velocities by a factor $\alpha \in \mathbb{R}$,
\[
v_m^{(\alpha)}(t) = \alpha\, v_m(t),
\]
yields a proportional scaling of the corresponding Doppler frequencies:
\[
f_m^{(\alpha)}(t) = \alpha\, f_m(t).
\]

This relationship is a direct consequence of 
Doppler frequency equation, making it a principled basis for probing whether 
learned models preserve the underlying velocity–frequency relationship.
\subsection{Frequency--Velocity Alignment (FVA)}
\label{sec:fva}
FVA evaluates the agreement between the model-predicted centroid trajectory and the reference trajectory specified by the RCS-Aware Doppler Model. Both sequences are defined over the same STFT frame index $n$, yielding $N = \lfloor (T - L)/H \rfloor + 1$ frames for a signal of length $T$. The metric is computed using the Pearson correlation:
\begin{equation}
\mathrm{FVA}
= \operatorname{corr}\!\left(
    f_c^{\mathrm{pred}}[n],\;
    f_c^{\mathrm{ref}}[n]
\right).
\label{eq:fva_def}
\end{equation}
The predicted centroid is derived from the ML model output. 
Given the model-predicted magnitude spectrogram 
$S^{\mathrm{pred}} \in \mathbb{R}^{N \times F}$ in dB scale, 
where $F$ is the number of frequency bins, the centroid for the 
$n$-th STFT frame is computed as the power-weighted mean:
\begin{equation}
\label{eq:centroid}
f_c^{\mathrm{pred}}[n]
= \frac{\sum_{k=0}^{F-1} P[n,k] \, f_k}
       {\sum_{k=0}^{F-1} P[n,k]},
\end{equation}
where $P[n,k] = 10^{S^{\mathrm{pred}}[n,k]/10}$ is the linear-scale 
power and $f_k$ is the $k$-th Doppler frequency bin. The frequency 
axis is defined over $[-f_s/2, f_s/2)$ with resolution $\Delta f = f_s/F$, 
so that $f_k = (k - F/2)\Delta f$.

\subsection{Doppler Consistency Score (DCS)}
\label{sec:dcs}
DCS evaluates whether the learned model preserves the expected velocity–frequency relationship under the velocity-scaling intervention defined in the Doppler Velocity-Frequency Model. For a set of applied scaling factors $\{\alpha_k\}$ and the corresponding model-predicted scaling factors $\{\alpha_{\mathrm{pred},k}\}$, DCS is 
defined as
\[
\mathrm{DCS}
= 1 -
\frac{\sum_{k=1}^{K}
    (\alpha_k - \alpha_{\mathrm{pred},k})^2}
     {\sum_{k=1}^{K} (\alpha_k - \bar{\alpha})^2},
\]where $\bar{\alpha} = \tfrac{1}{K}\sum_{k=1}^{K} \alpha_k$. A value of $\mathrm{DCS}=1$ indicates perfect proportionality ($\alpha_{\mathrm{pred},k}=\alpha_k$), $\mathrm{DCS}\approx 0$ corresponds to behavior comparable to a constant predictor, and negative values indicate worse-than-trivial scaling.

To obtain $\alpha_{\mathrm{pred}}$, we pass both the baseline 
($\alpha=1$) and $\alpha$-scaled MoCap through the model. From the resulting predicted spectrograms, we compute Doppler centroid sequences using~\eqref{eq:centroid}; let $f^{\mathrm{pred}}_{c, 1}[n]$ and $f^{\mathrm{pred}}_{c, \alpha}[n]$ denote the predicted centroids for the baseline and scaled MoCap inputs, respectively. The predicted scaling factor is then estimated via least-squares fitting:
\[
\alpha_{\mathrm{pred}}
= 
\frac{
\sum_{n=0}^{N-1} f^{\mathrm{pred}}_{c, \mathrm{1}}[n]\, f^{\mathrm{pred}}_{c, \alpha}[n]
}{
\sum_{n=0}^{N-1} \left(f^{\mathrm{pred}}_{c, \mathrm{1}}[n]\right)^2
}.
\]
\subsubsection*{Sign-flip variant}
We also introduce a sign-flip variant with $\alpha=-1$. Unlike the standard DCS, which evaluates linear proportionality, this variant focuses solely on whether the model preserves the correct Doppler sign reversal. Let $s^{\mathrm{exp}}[n] = \operatorname{sign}(-f^{\mathrm{pred}}_{c, \mathrm{1}}[n])$ denote the expected sign after velocity reversal, and $s^{\mathrm{pred}}[n] = \operatorname{sign}(f^{\mathrm{pred}}_{c, -1}[n])$ the predicted sign. The sign-consistency score is
\[
\mathrm{DCS}^{\mathrm{sign}}
= \frac{1}{2}\!\left(
1 + \frac{1}{N}\sum_{n=0}^{N-1}
s^{\mathrm{exp}}[n]\,
s^{\mathrm{pred}}[n]
\right),
\]
yielding a normalized score in $[0,1]$, where $1$ indicates perfect sign consistency and $0.5$ corresponds to random chance.
\subsection{Physical Interpretability via Metrics}
\label{sec:validity}

Together, MAE, FVA, and DCS provide a more complete picture of how the learned MoCap-to-radar model behaves. MAE measures reconstruction error (the training objective), while FVA and DCS assess physical consistency: FVA evaluates alignment with the physics-derived reference trajectory; DCS tests whether the model's Doppler response scales correctly with velocity, including proportional scaling and sign reversal.

Table~\ref{tab:metric_interpretation} illustrates how these metrics jointly characterize learned model behavior.

\begin{table}[h]
\centering
\caption{Joint interpretation of MAE, FVA, and DCS.}
\label{tab:metric_interpretation}
\begin{tabular}{ccc|l}
\toprule
MAE & FVA & DCS & Interpretation \\
\midrule
Low & High & High & Accurate and physically consistent \\
Low & Low  & Low  & Accurate but lacks physical grounding \\
High & High & High & Physically consistent but inaccurate \\
High & Low  & Low  & Neither accurate nor physically grounded \\
\bottomrule
\end{tabular}
\end{table}

\section{Experimental Setup}
\label{sec:exp-setup}
\subsection{Dataset and Preprocessing}

We use the MoCap2Radar dataset introduced in Chen et al.~\cite{chen2026mocap2radar}, collected in a calibrated $7 \times 7 \times 3.6$m MoCap lab using a  12-camera Vicon system and an Austere 5.8-GHz micro-Doppler radar from The Samraksh Company~\cite{roy2021one}. The Vicon system tracks $M = 53$ reflective markers at $f_m = 250$~Hz, while the radar provides complex baseband I/Q returns at $f_s = 256$~Hz. A synchronization signal from the Vicon controller is logged by both systems to ensure frame-level temporal alignment. Six reflective markers are attached to the radar housing so that its 3D position is recorded within the same global Vicon coordinate frame as the human subject. A single male adult participant performed both structured diagonal-walking trials at multiple speeds and unconstrained free-walking trials with spontaneous accelerations and decelerations, and frequent changes in walking direction. For model training, each structured trial is individually divided into training and validation segments. A single continuous free-walking trial is held out exclusively for testing. This split creates a generalization challenge, as the free-walking trial exhibits motion patterns that differ substantially from the 
directional and speed-controlled diagonal walks used for training.

After synchronizing both modalities to the radar sampling rate $f_s$, 
each trial is segmented into overlapping windows of length $L = 256$ 
samples with hop size $H = 32$.  An STFT with FFT size $F = 256$ and a Hann window is applied to the radar I/Q signal with DC removal. Each STFT window produces a single log-magnitude spectrum, and the 
sequence of spectra across time forms the trial-level spectrogram. 
The same window boundaries are applied to the aligned MoCap frames, 
yielding paired MoCap–spectrogram samples for supervised learning. 
All MoCap coordinates and log-magnitude spectra are standardized 
(z-score) using training-set statistics, with the same normalization 
applied to the validation and test sets. This preprocessing yields 
$N_{\text{train}} = 5757$ training samples, $N_{\text{val}} = 612$ 
validation samples, and $N_{\text{test}} = 2416$ held-out test samples. 
All evaluation metrics reported in this paper are computed exclusively 
on this held-out test set.

\subsection{MoCap2Radar Models for Interpretability Analysis}
We evaluate two categories of models: transformer-based architectures 
built on a spatial–temporal (ST) backbone, and a non-transformer
baseline.

\subsubsection*{Transformer-based Models}

\begin{figure}[!t]
    \centering
    \includegraphics[width=0.5\linewidth]{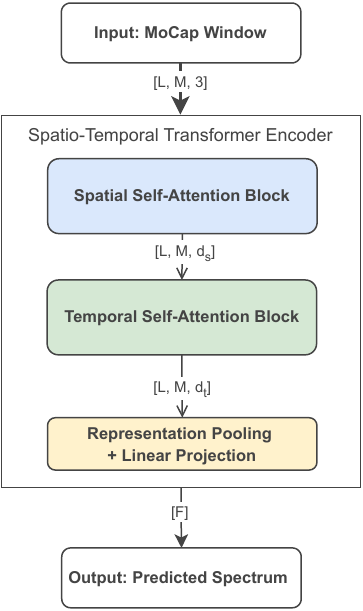}
    \caption{Overview of the MoCap2Radar backbone used in this work.}
    \label{fig:model}
\end{figure}

Fig.~\ref{fig:model} illustrates the spatio-temporal (ST) backbone 
used by the transformer models in this work. Each model takes a 
3D MoCap position window of shape $(L \times M \times 3)$ as input.  The backbone first applies a spatial self-attention block across the $M$ markers within each frame, followed by a temporal self-attention block across the $L$ frames. The output representation is then aggregated along the time axis and passed through a final linear projection to produce an $F$-dimensional magnitude spectrum. The models differ in either their spatial attention design or in whether specific attention components are ablated, as detailed below.

\textbf{ST (Flat).} This model follows the MoCap2Radar model introduced by Chen et al.~\cite{chen2026mocap2radar}. It applies a single spatial self-attention block over all $53$ markers within each frame, i.e., a flat spatial attention design, followed by a temporal self-attention block over the $L = 256$ frames. Its non-hierarchical formulation serves as the reference architecture.

\textbf{HST (Full).} We propose this model to serve as a physically motivated variant in our interpretability study, incorporating a hierarchical two-stage spatial attention mechanism. Its design is inspired by the anatomical organization of the Plug-in Gait marker set used in the Vicon MoCap system~\cite{davis1991gait}. Markers are grouped into anatomically coherent body segments, enabling the model to perform intra-segment attention followed by inter-segment attention before temporal processing. The hierarchy imposes a physically meaningful spatial inductive bias, encouraging the model to learn motion relationships that are consistent with human anatomy. 

To analyze the contribution of different attention components, we derive three ablation variants from HST (Full). These variants disable specific attention paths while keeping all learned parameters and the overall architecture unchanged, isolating the functional role of each attention component under controlled conditions.

\textbf{HST (No-Attn).}
Disables both spatial and temporal attention. The model instead uses mean pooling across markers and across frames at the corresponding stages.

\textbf{HST (No-Spatial).}
Disables the hierarchical spatial attention blocks while retaining the temporal attention module. Spatial information is aggregated via per-frame mean pooling.

\textbf{HST (No-Temporal).}
Disables the temporal attention module while retaining the hierarchical spatial attention blocks. Temporal aggregation is performed using mean pooling.

\subsubsection*{Non-transformer Baseline}

\textbf{MLP (Naive).} Unlike the transformer models above, this 
baseline has no explicit spatial–temporal structure. It flattens 
the input window and predicts the output spectrogram using a single 
hidden layer (512 units, ReLU activation). It serves as a lower 
bound to test whether structured, transformer-based modeling is 
necessary for the MoCap-to-radar mapping.

\subsection{Training Configuration}

The architectural hyperparameters used for all models are listed in Table~\ref{tab:hyperparameters}. All models are trained under an identical optimization protocol using AdamW with an initial learning rate of $1\times10^{-4}$ and a batch size of 32. Training uses the MAE loss on the normalized log-magnitude spectrograms. A ReduceLROnPlateau scheduler is applied with a reduction factor of 0.5 and a patience of 10 epochs. Early stopping is used with a patience of 40 epochs, 
and training proceeds for up to 200 epochs. All experiments are replicated with six random seeds, and the checkpoint achieving the lowest validation MAE loss is used for test evaluation. 

\begin{table}[t]
\centering
\caption{Model architecture hyperparameters.}
\label{tab:hyperparameters}
\begin{tabular}{ll}
\toprule
\textbf{Parameter} & \textbf{Value} \\
\midrule
Spatial embedding dimension ($d_s$) & 128 \\
Temporal embedding dimension ($d_t$) & 256 \\
Feedforward width            & 512 \\
Spatial attention heads      & 8 \\
Temporal attention heads     & 8 \\
Dropout rate           & 0.1 \\
\bottomrule
\end{tabular}
\end{table}

\section{Experimental Results and Discussion}
\label{sec:exp-results}
\begin{figure}[t]
    \centering
\caption{
Ground-truth spectrogram (30s) with the physics-derived Doppler centroid, shown alongside the predicted spectrograms and centroid trajectories from HST~(Full), HST~(No-Temporal), and MLP~(Naive).
}
    \includegraphics[width=\linewidth]{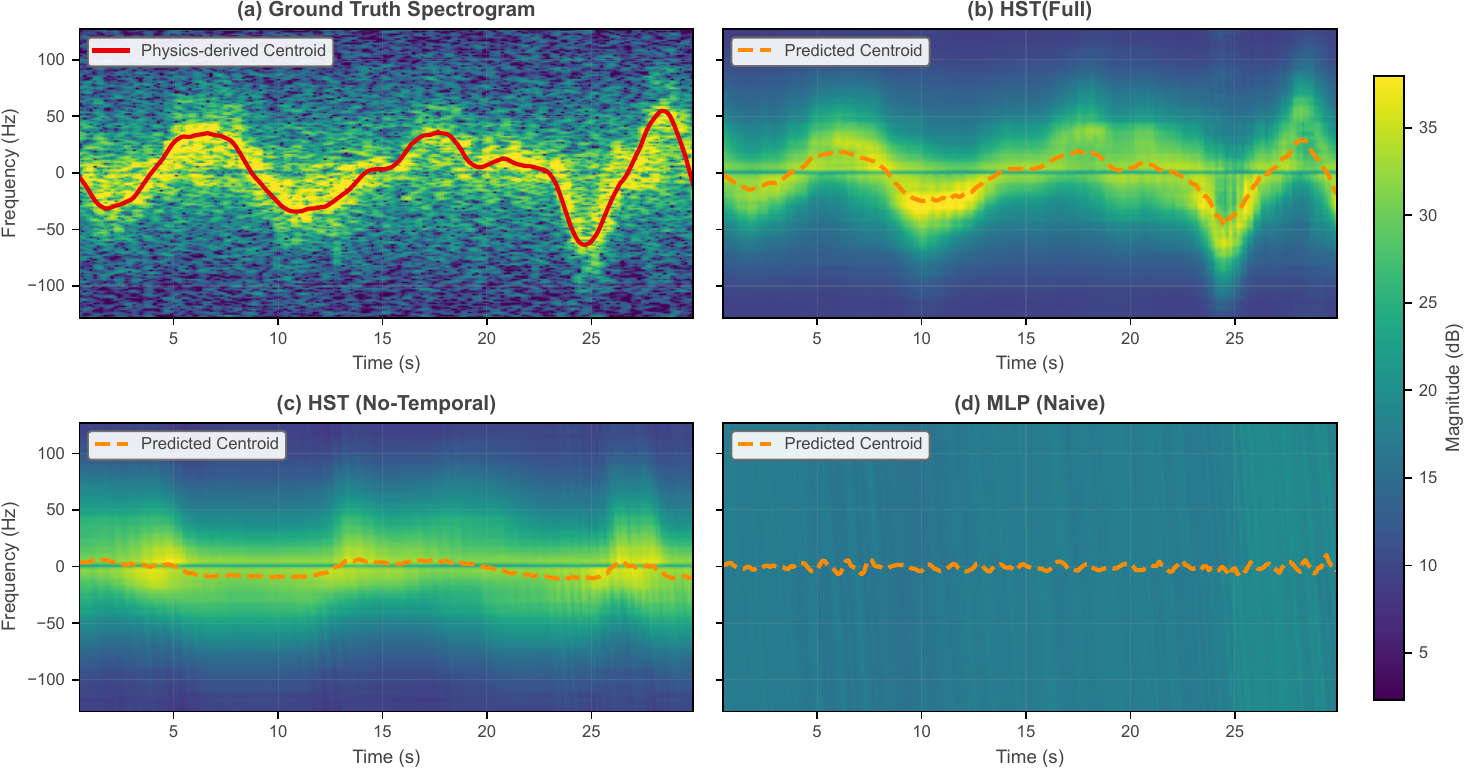}
    \label{fig:randomwalk_comparison_4panels}
\end{figure}
\subsection{Analysis of MAE and FVA Performance}

\begin{table*}[t]
\centering
\caption{Performance metrics for all model variants, reported as mean~$\pm$~standard deviation over six random seeds.}
\label{tab:performance}
\begin{tabular}{lcccccc}
\toprule
\textbf{Model} 
& \textbf{MAE$_\text{dB}$ $\downarrow$} 
& \textbf{FVA$\uparrow$} 
& \textbf{DCS $\uparrow$}
& \textbf{FVA$^{\text{rev}}$ $\uparrow$} 
& \textbf{DCS$^{\text{sign}}$ $\uparrow$} 
\\
\midrule
MLP (Naive)            
& $8.034 \pm 0.019$ 
& $0.010 \pm 0.036$ 
& $0.654 \pm 0.217$
& $-0.005 \pm 0.041$ 
& $ 0.305 \pm 0.039 $ 
 
\\
ST (Flat)              
& $4.760 \pm 0.119$ 
& $0.945 \pm 0.005$ 
& $0.969 \pm 0.020$
& $0.660 \pm 0.065$ 
& $0.695 \pm 0.023$ 
\\
HST (Full)             
& $4.823 \pm 0.075$ 
& $0.928 \pm 0.004$ 
& $0.918 \pm 0.050$
& $0.672 \pm 0.024$ 
& $0.677 \pm 0.015 $ 
\\
HST (No-Attn)          
& $4.997 \pm 0.062$ 
& $0.279 \pm 0.032$ 
& $-0.091 \pm 1.101$
& $-0.284 \pm 0.008$ 
& $ 0.062 \pm 0.016$ 
\\
HST (No-Spatial)     
& $4.725 \pm 0.064$  
& $0.944 \pm 0.006$ 
& $0.974 \pm 0.011$
& $0.634 \pm 0.009$ 
& $0.697 \pm 0.011  $ 
\\
HST (No-Temporal)       
& $4.947  \pm 0.042$ 
& $0.302 \pm 0.025$ 
& $-1.116 \pm 0.450$
& $-0.268 \pm 0.032$ 
& $ 0.051 \pm 0.016 $ 
\\
\bottomrule
\end{tabular}
\end{table*}

Temporal attention is the primary factor driving high FVA. Table~\ref{tab:performance} summarizes the metrics in our framework. As expected, the MLP baseline performs the worst in both MAE and FVA, and its near-zero FVA indicates that it fails to learn any meaningful alignment between the predicted and reference Doppler centroid trajectories. The remaining models separate naturally into two groups based on the presence or absence of temporal attention. Models with temporal attention achieve substantially higher FVA. This can be interpreted as a consequence of Doppler frequencies being determined by the time-varying radial velocity, which makes Doppler signatures strongly dependent on temporal evolution. A model lacking temporal attention cannot aggregate frame-to-frame MoCap information and thus exhibits limited physical coherence. 

Models with comparable MAE can nonetheless differ substantially in 
their FVA. Between models with and without temporal attention, MAE 
differs by less than $0.3$~dB ($\sim$ 6\%), yet FVA differs 
by over $0.6$ ($\sim$ 65\%). This suggests that magnitude-based 
errors are not sensitive to Doppler centroid trajectory deviations, 
whereas FVA captures alignment differences not apparent from MAE alone. Furthermore, we observe a mild divergence between MAE and FVA rankings: ST (Flat) obtains the highest FVA ($0.945$) with the second-best MAE ($4.760$~dB), whereas HST (No-Spatial) achieves the lowest MAE ($4.725$~dB) but the second-highest FVA ($0.944$). Although these differences are rather small, the reversed ordering of the two models under MAE and FVA suggests that the metrics may capture distinct aspects of the prediction task and may emphasize different properties of the learned mapping. Determining whether this reflects a consistent trade-off or merely dataset-level variability warrants further investigation.

Finally, we note that HST (Full) does not outperform the simpler 
model variants, despite incorporating a hierarchical spatial-attention 
mechanism. Although the design was intended to encode structured 
spatial relationships among body parts, its empirical benefit is 
not evident in our current setting. Because the performance 
differences are small, we avoid attributing this outcome to any 
specific modeling factor. At a minimum, these results indicate 
that the present spatial hierarchy does not consistently improve 
performance. Whether more refined spatial modeling could offer 
measurable advantages, or whether the coarse RCS approximation in 
our physical reference limits the ability to detect such 
improvements, remains an open question for future work.

\subsection{Qualitative Comparison via Spectrograms}
The qualitative comparison reveals that the visual quality of the 
predicted spectrograms correlates closely with their FVA scores. 
Fig.~\ref{fig:randomwalk_comparison_4panels} provides a qualitative 
comparison of the first 30\,s of the held-out test sequence. In 
Fig.~\ref{fig:randomwalk_comparison_4panels}\,(a), the alignment 
between the physics-derived reference centroid and the dominant 
spectral trend in the ground-truth spectrogram is visually apparent. 
Fig.~\ref{fig:randomwalk_comparison_4panels}\,(b)--(d) show the 
predictions from HST (Full), HST (No-Temporal), and MLP (Naive), 
respectively. As can be seen, HST (Full) captures the overall micro-Doppler pattern, and the centroid trajectory computed from its 
predicted spectrogram follows the reference reasonably well, 
consistent with its low MAE and high FVA. HST (No-Temporal) produces 
predicted spectrograms with visible temporal inconsistencies and a 
noticeably flattened trajectory. These effects lead to the large 
drop in FVA, and the magnitude of this drop is consistent with the 
substantial visual distortion observed both in the predicted 
spectrogram and in the predicted centroid trajectory. In contrast, 
the MAE worsens only slightly and provides little indication of 
the severe distortions present, indicating that FVA reliably 
reflects perceptible degradation in Doppler centroid 
that MAE alone cannot capture. MLP (Naive) produces a severely 
distorted and physically implausible prediction, failing to capture 
any coherent micro-Doppler features. Its high MAE and near-zero FVA 
accurately reflect this failure.
\subsection{Velocity-Scaling Consistency Analysis}
DCS further highlights the role of temporal attention in preserving 
the velocity–frequency relationship. We evaluate velocity scaling 
using $K=10$ scaling factors uniformly sampled from $[0,1]$, i.e., 
$\alpha \in \{0.1,0.2,\dots,1.0\}$, selected to keep the induced 
Doppler shifts within the radar's operational bandwidth. As shown 
in Table~\ref{tab:performance}, DCS follows the same overall trend 
observed in MAE and FVA. Interestingly, the MLP (Naive) attains 
moderate DCS scores, even substantially outperforming transformer variants in 
which temporal attention has been ablated. However, this does not necessarily imply that 
the MLP learns the underlying physics; rather, a plausible 
explanation is that a simple MLP behaves approximately as a global 
near-linear mapping, so scaling the input motion by $\alpha$ tends 
to produce a proportionally scaled centroid prediction. In contrast, 
a transformer without temporal attention lacks any sequence-level 
mechanism to enforce global scaling. Its feedforward, residual, and 
normalization operations act locally in time and introduce additional 
nonlinearities, leading to inconsistent responses under velocity 
scaling and degraded DCS scores.

We also examine two additional metrics, FVA$^{\mathrm{rev}}$ and 
DCS$^{\mathrm{sign}}$, corresponding to the two rightmost columns of 
Table~\ref{tab:performance}. The FVA$^{\mathrm{rev}}$ score is 
obtained by applying the same FVA definition~\eqref{eq:fva_def}
to the reference trajectory computed from the velocity-reversed 
MoCap sequence. Its trends mirror those of standard FVA, although 
all models exhibit reduced scores due to the altered spectral 
structure under sign reversal. For DCS$^{\mathrm{sign}}$, only models 
with temporal attention maintain performance above 0.5. Models 
without temporal attention collapse to chance-level behavior. The 
naive MLP baseline performs slightly better than these temporal-ablated 
transformer variants but still remains below 0.5, consistent with 
the near-linearity explanation and confirming that it does not 
reliably preserve sign consistency under velocity reversal.

\subsection{Limitations and Future Work}

The proposed FVA and DCS metrics provide necessary but not sufficient conditions for physical consistency. Low scores reliably indicate a failure to capture the underlying physics, but high scores confirm agreement only with the specific physics models employed and do not guarantee broader physical correctness. In addition, FVA and DCS probe model behavior only at the output level. They do not reveal the internal mechanisms responsible for physical consistency, and understanding these learned representations remains an open direction for future work. The present framework also focuses on centroid-level physical consistency; extending it to other spectral attributes (e.g., bandwidth, harmonic structure, micro-motion periodicity) offers another promising avenue for future investigation. Finally, the dataset comprises a single participant, and broader validation across diverse subjects and body types remains a direction for future work.

\section{Conclusion}
\label{sec:conclusion}
Understanding what physics a model has learned has implications for 
its trustworthiness, generalizability, and characterization. By probing 
data-driven MoCap-to-radar models through the proposed framework, we reveal that 
models with similar reconstruction error can differ substantially in 
physical consistency. Beyond evaluation, the framework also helps 
pinpoint which architectural components are essential for capturing 
the underlying physics. Together, these findings suggest that 
physics-grounded metrics offer insights into ML-based model behavior that 
the standard reconstruction error alone cannot provide.

Looking forward, this work points toward a broader paradigm: 
systematically probing learned models against known physical laws 
to assess not just predictive accuracy, but physical validity. 
Although demonstrated in this work for MoCap-to-radar synthesis, the principle 
generalizes: wherever physical laws govern a data-driven generative 
model's output, those laws can serve as interpretability benchmarks, 
offering a path toward generative models that are not only accurate 
but verifiably grounded in the physics of their domain.
\bibliographystyle{IEEEtran}
\bibliography{ref}

\end{document}